\documentclass[letterpaper]{article} 
\usepackage{aaai23}  
\usepackage{times}  
\usepackage{helvet}  
\usepackage{courier}  
\usepackage[hyphens]{url}  
\usepackage{graphicx} 
\usepackage{natbib}  
\usepackage{caption} 
\usepackage{algorithm}
\usepackage{algorithmic}

\usepackage{multirow}
\usepackage{amsmath}
\usepackage{amssymb}
\usepackage{booktabs}

\usepackage{newfloat}
\usepackage{listings}
\DeclareCaptionStyle{ruled}{labelfont=normalfont,labelsep=colon,strut=off} 
\floatstyle{ruled}
\newfloat{listing}{tb}{lst}{}
\floatname{listing}{Listing}
\title{Temporal Knowledge Graph Reasoning with Historical Contrastive Learning}
\author{
    Yi Xu,
    Junjie Ou,
    Hui Xu,
    Luoyi Fu\thanks{Corresponding author.}
}
\affiliations{

    Department of Computer Science and Engineering\\
    Shanghai Jiao Tong University\\
    \{yixu98, j\_michael, xhui\_1, yiluofu\}@sjtu.edu.cn
}

\begin{document}

\maketitle

\begin{abstract}
Temporal knowledge graph, serving as an effective way to store and model dynamic relations, shows promising prospects in event forecasting. However, most temporal knowledge graph reasoning methods are highly dependent on the recurrence or periodicity of events, which brings challenges to inferring future events related to entities that lack historical interaction. In fact, the current moment is often the combined effect of a small part of historical information and those unobserved underlying factors. To this end, we propose a new event forecasting model called \textbf{C}ontrastive \textbf{E}vent \textbf{Net}work (CENET), based on a novel training framework of historical contrastive learning. CENET learns both the historical and non-historical dependency to distinguish the most potential entities that can best match the given query. Simultaneously, it trains representations of queries to investigate whether the current moment depends more on historical or non-historical events by launching contrastive learning. The representations further help train a binary classifier whose output is a boolean mask to indicate related entities in the search space. During the inference process, CENET employs a mask-based strategy to generate the final results. We evaluate our proposed model on five benchmark graphs. The results demonstrate that CENET significantly outperforms all existing methods in most metrics, achieving at least $8.3\%$ relative improvement of Hits@1 over previous state-of-the-art baselines on event-based datasets.
\end{abstract}

\section{Introduction}
\begin{figure}[t]
    \centering\textbf{}
    \includegraphics[width=1.0\linewidth]{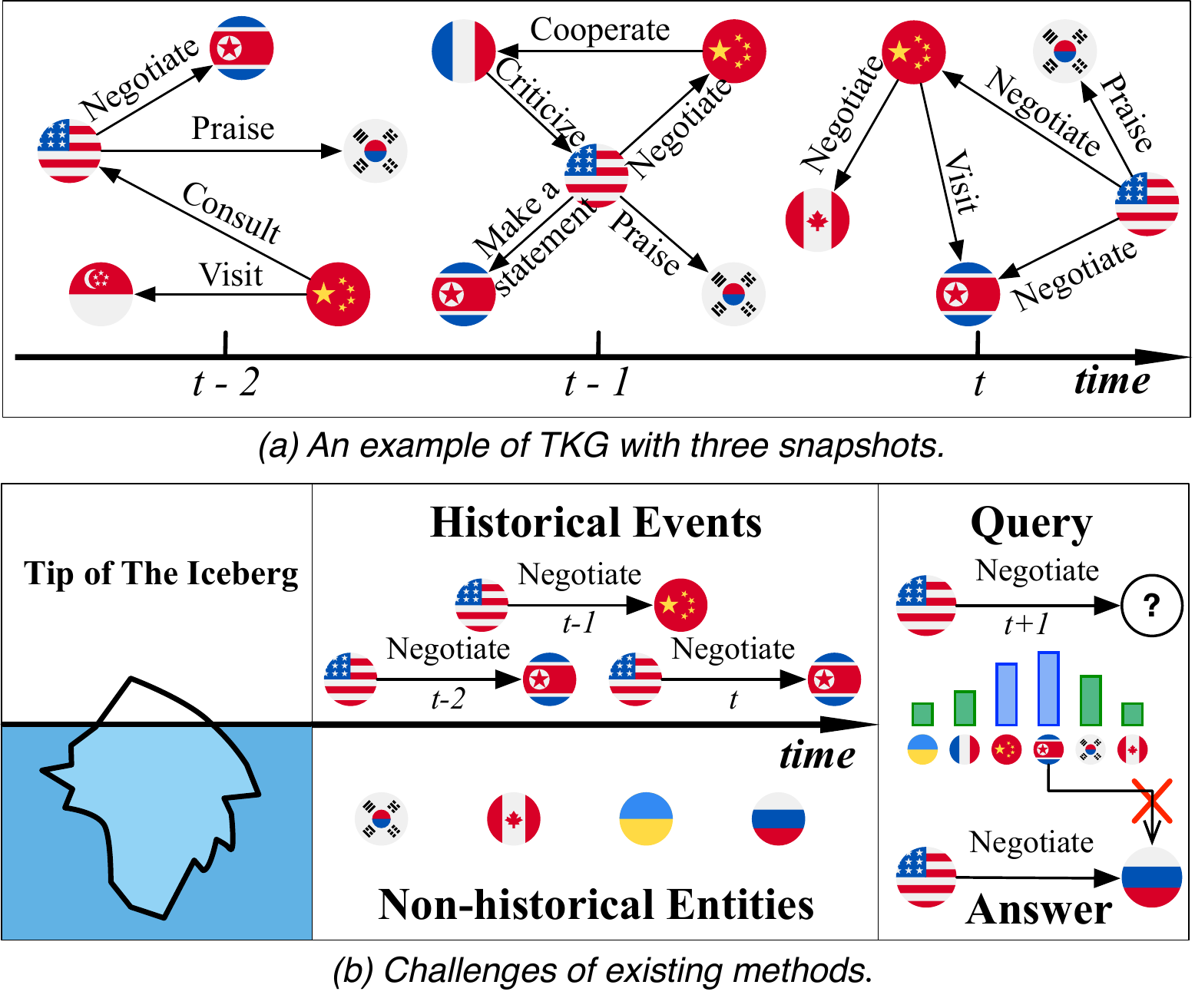}
    \caption{An example of TKG and challenges of existing methods.}
    \label{fig:tkg}
\end{figure}

Knowledge Graphs (KGs), serving as the collections of human knowledge, have revealed promising expectations in the field of natural language processing~\cite{sun2020colake, wang2020covid}, recommendation system~\cite{wang2019kgat}, and information retrieval~\cite{liu2018entity}, etc. A traditional KG is usually a static knowledge base that uses a graph-structured data topology to integrate facts (also called events) in the form of triples $(s, p, o)$, where $s$ and $o$ denote subject and object entities respectively, and $p$ as a relation type means predicate. In the real world, knowledge evolves continuously, inspiring the construction and application of the Temporal Knowledge Graphs (TKGs), where the fact has extended from a triple $(s, p, o)$ to a quadruple with a timestamp $t$, i.e., $(s, p, o, t)$. As a result, a TKG consists of multiple snapshots, and the facts in the same snapshot co-occur. Figure~\ref{fig:tkg} (a) shows an example of TKG consisting of a series of international political events, where some events may occur repeatedly, and new events will also emerge.


TKGs provide new perspectives and insights for many downstream applications, e.g., policymaking~\cite{deng2020dynamic}, stock prediction~\cite{feng2019temporal}, and dialogue systems~\cite{jia2018tequila}, thus triggering intense interests in TKG reasoning. In this work, we focus on forecasting events (facts) in the future on TKGs, which is also called graph extrapolation. Our goal is to predict the missing entities of queries like $(s, p, ?, t)$ for a future timestamp $t$ that has not been observed in the training set.


Many efforts~\cite{garcia2018learning, jin2020recurrent} have been made toward modeling the structural and temporal characteristics of TKGs for future event prediction. Some mainstream examples~\cite{jin2020recurrent, li2021temporal} make reference to known events in history, which can easily predict repetitive or periodic events. However, in terms of the event-based TKG \textit{Integrated Crisis Early Warning System}, new events that have never occurred before account for about $40\%$  \cite{boschee2015icews}. It is challenging to infer these new events because they have fewer temporal interaction traces during the whole timeline. For instance, the right part of Figure~\ref{fig:tkg} (b) shows the query \textit{(the United States, Negotiate, ?, t+1)} and its corresponding new events \textit{(the United States, Negotiate, Russia, t+1)}, where most existing methods often obtain incorrect results over such query due to their focus on the high frequent recurring events.
Additionally, during the inference process, existing methods rank the probability scores of overall candidate entities in the whole graph without any bias. We argue that the bias is necessary when approaching the missing entities of different events. For repetitive or periodic events, models are expected to prioritize a few frequently occurring entities, and for new events, models should pay more attention to entities with less historical interaction.




In this work, we will go beyond the limits of historical information and mine potential temporal patterns from the whole knowledge. To elaborate our design clearer, we call the past events associated with the entities in the current query $(s, p, ?, t)$ \textit{historical events}, and others \textit{non-historical events}. Their corresponding entities are called \textit{historical} and \textit{non-historical entities}, respectively. We will give formal definitions in Section~\ref{preliminaries}. We intuitively consider that the events in TKG are not only related to their historical events but also indirectly related to unobserved underlying factors. The historical events we can see are only the tip of the iceberg. We propose a novel TKG reasoning model called CENET (\textbf{C}ontrastive \textbf{E}vent \textbf{Net}work) for event forecasting based on contrastive learning. Given a query $(s, p, ?, t)$ whose real object entity is $o$, CNENT takes into account its historical and non-historical events and identify significant entities via contrastive learning. Specifically, a copy mechanism-based scoring strategy is first adopted to model the dependency of historical and non-historical events. In addition, all queries can be divided into two classes according to their real object entities: either the object entity $o$ is a historical entity or a non-historical entity. Therefore, CENET naturally employs supervised contrastive learning to train representations of the two classes of queries, further helping train a classifier whose output is a boolean value to identify which kind of entities should receive more attention. During the inference, CENET combines the distribution from the historical and non-historical dependency, and further considers highly correlated entities with a mask-based strategy according to the classification results.


The contributions of our paper are summarized as follows:

\begin{itemize}
    \item We propose a TKG model called CENET for event forecasting. CENET can predict not only repetitive and periodic events but also potential new events via joint investigation of both historical and non-historical information;
    
    \item To the best of our knowledge, CENET is the first model to apply contrastive learning to TKG reasoning, which trains contrastive representations of queries to identify highly correlated entities;
    
    \item We conduct experiments on five public benchmark graphs. The results demonstrate that CENET outperforms the state-of-the-art TKG models in the task of event forecasting.
\end{itemize}


\section{Related Work}\label{sec:related}

\subsection{Temporal Knowledge Graph Reasoning}
There are two different settings for TKG reasoning: interpolation and extrapolation~\cite{jin2020recurrent}. Given a TKG with timestamps ranging from $t_0$ to $t_n$, models with the interpolation setting aim to complete missing events that happened in the interval $[t_0, t_n]$, which is also called TKG completion. In contrast, the extrapolation setting aims to predict possible events after the given time $t_n$, i.e., inferring the entity $o$ (or $s$) given query $q=(s,p,?,t)$ (or $(?,p,o,t)$) where $t > t_n$. 

Models in the former case such as HyTE~\cite{dasgupta2018hyte}, TeMP~\cite{wu2020temp}, and ChronoR~\cite{sadeghian2021chronor} are designed to infer missing relations within the observed data. However, such models are not designed to predict future events that fall out of the specified time interval. In the latter case, various methods are designed for the purpose of future event prediction. Know-Evolve~\cite{trivedi2017know} is the first model to learn non-linearly evolving entity embeddings, yet unable to capture the long-term dependency. xERTE~\cite{han2020explainable} and TLogic~\cite{liu2021tlogic} provide understandable evidence that can explain the forecast, but their application scenarios are limited. TANGO~\cite{han2021learning} employs neural ordinary differential equations to model the TKGs. A copy-generation mechanism is adopted in CyGNet~\cite{zhu2021learning} to identify high-frequency repetitive events. CluSTeR~\cite{li2021search} is designed with reinforcement learning, yet constraining its applicability to event-based TKGs. There also emerge some models which try to adopt GNN~\cite{kipf2016semi} or RNN architecture to capture spatial temporal patterns. Typical examples include RE-NET~\cite{jin2020recurrent}, RE-GCN~\cite{li2021temporal}, HIP~\cite{he2021hip}, and EvoKG~\cite{park2022evokg}.



\begin{figure*}[ht]
    \centering
    \includegraphics[width=1.0\linewidth]{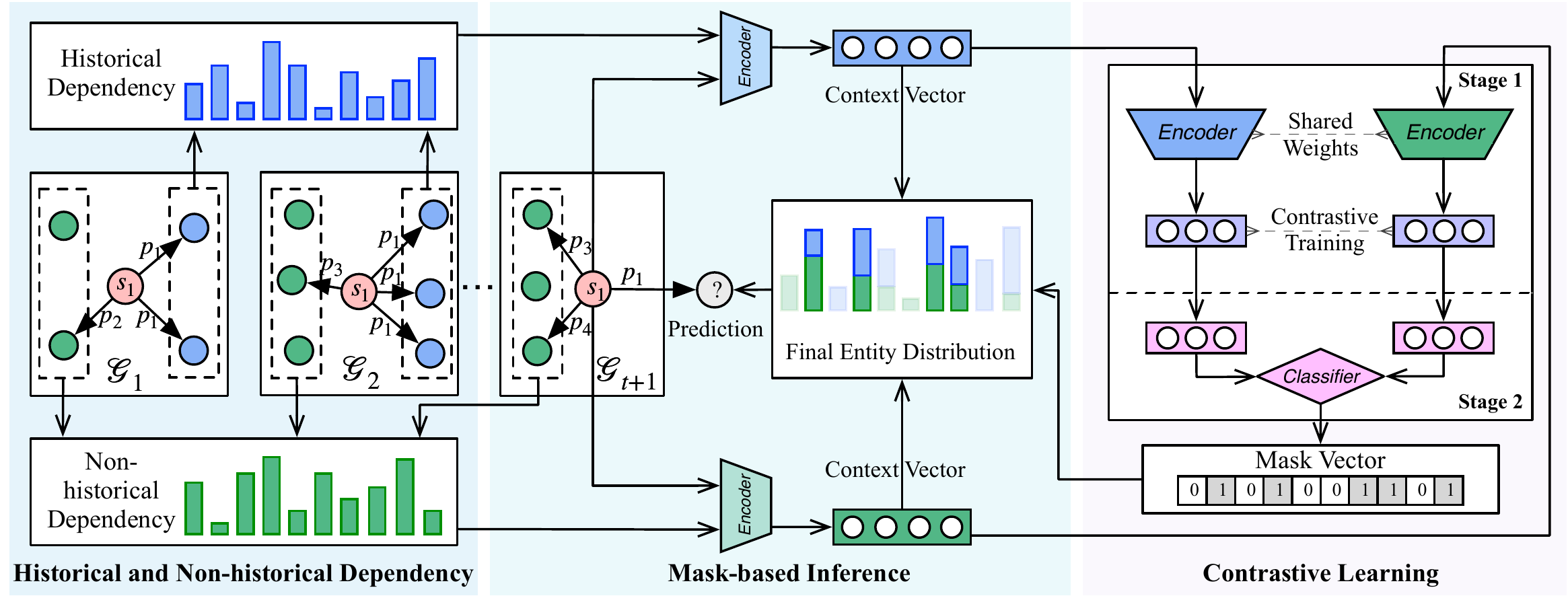}
    \caption{The overall architecture of CENET. The left part learns the distribution of entities from both historical and non-historical dependency. The right part illustrates the two stages of historical contrastive learning, which aims to identify highly correlated entities, and the output is a boolean mask vector. The middle part is the mask-based inference process that combines the distribution learned from the two kinds of dependency and the mask vector to generate the final results.}
    \label{fig:architecture}
\end{figure*}

\subsection{Contrastive Learning}
Contrastive learning as a self-supervised learning paradigm focuses on distinguishing instances of different categories. In self-supervised contrastive learning, most methods~\cite{chen2020simple} derive augmented examples from a randomly sampled minibatch of $N$ examples, resulting in $2N$ samples to optimize the following loss function given a positive pair of examples $(i,j)$. Equation~\ref{eq:contrastive_loss} is the contrastive loss:

\begin{equation}\label{eq:contrastive_loss}
    \mathcal{L}_{i,j} = - \log \frac{exp(\textbf{z}_i \cdot \textbf{z}_j / \tau)}{\sum_{k=1,k\neq i}^{2N}exp(\textbf{z}_i \cdot \textbf{z}_k / \tau)},
\end{equation}
where $\textbf{z}_i$ is the projection embedding of sample $i$ and $\tau \in \mathbb{R}^{+}$ denotes a temperature parameter helping the model learn from hard negatives. In the case of supervised learning, there is a work~\cite{khosla2020supervised} generalizing contrastive loss to an arbitrary number of positives, which separates the representations of different instances using ground truth labels. The obtained contrastive representations can promote the downstream classifier to achieve better performance compared with vanilla classification model. 



\section{Method}\label{method}
As shown in Figure~\ref{fig:architecture}, CENET captures both the historical and non-historical dependency. Simultaneously, it utilizes contrastive learning to identify highly correlated entities. A mask-based inference process is further employed for reasoning performing. In the following parts, we will introduce our proposed method in detail.

\subsection{Preliminaries}\label{preliminaries}
Let $\mathcal{E}$, $\mathcal{R}$, and $\mathcal{T}$ denote a finite set of entities, relation types, and timestamps, respectively. A temporal knowledge graph $\mathcal{G}$ is a set of quadruples formalized as $(s,p,o,t)$, where $s \in \mathcal{E}$ is a subject (head) entity, $o \in \mathcal{E}$ is an object (tail) entity, $p \in \mathcal{R}$ is the relation (predicate) occurring at timestamp $t$ between $s$ and $o$. $\mathcal{G}_t$ represents a TKG snapshot which is the set of quadruples occurring at time $t$. We use boldfaced $\textbf{s}, \textbf{p}, \textbf{o}$ for the embedding vectors of $s$, $p$, and $o$ respectively, the dimension of which is $d$. $\textbf{E} \in \mathbb{R}^{|\mathcal{E}| \times d}$ is the embeddings of all entities, the row of which represents the embedding vector of an entity such as $\textbf{s}$ and $\textbf{o}$. Similarly, $\textbf{P} \in \mathbb{R}^{|\mathcal{R}| \times d}$ is the embeddings of all relation types.

Given a query $q=(s,p,?,t)$, we define the set of \textit{historical events} as $\mathcal{D}_{t}^{s,p}$ and the corresponding set of \textit{historical entities} as $\mathcal{H}_{t}^{s,p}$ in the following equations:
\begin{equation}\label{eq:historical_event}
   \mathcal{D}_{t}^{s,p}=\bigcup_{k < t}\{(s,p,o,k) \in \mathcal{G}_k\},
\end{equation}

\begin{equation}\label{eq:historical_entity}
    \mathcal{H}_{t}^{s,p}=\{o|(s,p,o,k) \in \mathcal{D}_{t}^{s,p}\}.
\end{equation}
Naturally, entities not in $\mathcal{H}_{t}^{s,p}$ are called \textit{non-historical entities}, and the set $\{(s,p,o',k)| o' \not\in \mathcal{H}_{t}^{s,p}, k < t\}$ denotes the set of \textit{non-historical events}, where some quadruples may not exist in $\mathcal{G}$. It is worth noting that we also use $\mathcal{D}_{t}^{s,p}$ to represent the set of historical events for a current event $(s, p, o, t)$. If an event $(s,p,o,t)$ itself does not exist in its corresponding $\mathcal{D}_{t}^{s,p}$, then it is a new event. Without loss of generality, we detail how CENET predicts object entities with a given query $q=(s,p,?,t)$ in the following parts.



\subsection{Historical and Non-historical Dependency}
In most TKGs, although many events often show repeated occurrence pattern, new events may have no historical events to refer to. To this end, CENET takes not only historical but also non-historical entities into consideration. We first investigate the frequencies of historical entities for the given query $q=(s,p,?,t)$ during data pre-processing. More specifically, we count the frequencies $\textbf{F}_{t}^{s,p} \in \mathbb{R}^{|\mathcal{E}|}$ of all entities served as the objects associated with subject $s$ and predicate $p$ before time $t$, as shown in Equation~\ref{eq:frequency}:

\begin{equation}\label{eq:frequency}
    \textbf{F}_{t}^{s,p}(o)=\sum_{k < t}|\{o|(s,p,o,k) \in \mathcal{G}_k\}|.
\end{equation}
Since we cannot count the frequencies of non-historical entities, CENET transforms $\textbf{F}_{t}^{s,p}$ into $\textbf{Z}_{t}^{s,p} \in \mathbb{R}^{|\mathcal{E}|}$ where the value of each slot is limited by a hyper-parameter $\lambda$:

\begin{equation}\label{eq:frequency_trans}
    \textbf{Z}_{t}^{s,p}(o) = \lambda \cdot (\Phi_{\textbf{F}_{t}^{s,p}(o) > 0} - \Phi_{\textbf{F}_{t}^{s,p}(o) = 0}).
\end{equation}
$\Phi_{\beta}$ is an indicator function that returns 1 if $\beta$ is true and 0 otherwise. $\textbf{Z}_{t}^{s,p}(o) > 0$ represents the quadruple $(s,p,o,t_k)$ is a historical event bound to $s, p,$ and $t$ $(t_k < t)$, while $\textbf{Z}_{t}^{s,p}(o) < 0$ indicates that the quadruple $(s,p,o,t_k)$ is a non-historical event that does not exist in $\mathcal{G}$. Next, CENET learns the dependency from both the historical and non-historical events based on the input $\textbf{Z}_{t}^{s,p}$. CENET adopts a copy mechanism based learning strategy~\cite{gu2016incorporating} to capture different kinds of dependency from two aspects: one is the similarity score vector between query and the set of entities, the other is the query's corresponding frequency information with copy mechanism.

For historical dependency, CENET generates a latent context vector $\textbf{H}_{his}^{s,p} \in \mathbb{R}^{|\mathcal{E}|}$ for query $q$, which scores the historical dependency of different object entities:

\begin{equation}\label{eq:hhis}
    \textbf{H}_{his}^{s,p}=\underbrace{tanh(\textbf{W}_{his}(\textbf{s}\oplus \textbf{p}) + \textbf{b}_{his})\textbf{E}^{T}}_{\text{similarity score between $q$ and $\mathcal{E}$}} + \textbf{Z}_{t}^{s,p},
\end{equation}
where \textit{tanh} is the activation function, $\oplus$ represents the concatenation operator, $\textbf{W}_{his} \in \mathbb{R}^{d \times 2d}$ and $\textbf{b}_{his} \in \mathbb{R}^{d}$ are trainable parameters. We use a linear layer with \textit{tanh} activation to aggregate the query's information. The output of the linear layer is then multiplied by \textbf{E} to obtain an $|\mathcal{E}|$-dimensional vector, where each element represents the \textbf{similarity} score between the corresponding entity $o' \in \mathcal{E}$ and the query $q$. Then, according to the copy mechanism, we add the copy-term $\textbf{Z}_{t}^{s,p}$ to change the index scores of historical entities in $\textbf{H}_{his}^{s,p}$ to higher values directly without contributing to the gradient update. Thus, $\textbf{Z}_{t}^{s,p}$ makes $\textbf{H}_{his}^{s,p}$ pay more attention to historical entities. Similarly, for non-historical dependency, the latent context vector $\textbf{H}_{nhis}^{s,p}$ is defined as:

\begin{equation}\label{eq:hnhis}
    \textbf{H}_{nhis}^{s,p}=tanh(\textbf{W}_{nhis}(\textbf{s}\oplus \textbf{p}) + \textbf{b}_{nhis})\textbf{E}^{T} - \textbf{Z}_{t}^{s,p}.
\end{equation}
Contrary to historical dependency (Equation~\ref{eq:hhis}), subtracting $\textbf{Z}_{t}^{s,p}$ makes $\textbf{H}_{nhis}^{s,p}$ focus on non-historical entities. The training objective of learning from both historical and non-historical events is to minimize the following loss $\mathcal{L}^{ce}$:



\begin{equation}\footnotesize\label{eq:ce_loss}
    \mathcal{L}^{ce} = -\sum_{q}\log\{\frac{exp(\textbf{H}_{his}^{s,p}(o_i))}{\sum\limits_{o_j \in \mathcal{E}}exp(\textbf{H}_{his}^{s,p}(o_j))} + \frac{exp(\textbf{H}_{nhis}^{s,p}(o_i))}{\sum\limits_{o_j \in \mathcal{E}}exp(\textbf{H}_{nhis}^{s,p}(o_j))}\},
\end{equation}
where $o_i$ denotes the ground truth object entity of the given query $q$. The purpose of $\mathcal{L}^{ce}$ is to separate ground truth from others by comparing each scalar value in $\textbf{H}_{his}^{s,p}$ and $\textbf{H}_{nhis}^{s,p}$.



During the inference, CENET combines the softmax results of the above two latent context vectors as the predicted probabilities $\textbf{P}^{s,p}_{t}$ over all object entities:
\begin{equation}\label{eq:prob}
    \textbf{P}^{s,p}_{t}=\frac{1}{2}\{softmax(\textbf{H}_{his}^{s,p}) + softmax(\textbf{H}_{nhis}^{s,p})\},
\end{equation}
where the entity with maximum value is the most likely entity the component predicts. 


\subsection{Historical Contrastive Learning}
Clearly, the learning mechanism defined above well captures the historical and non-historical dependency for each query. However, many repetitive and periodic events are only associated with historical entities. Besides, for new events, existing models are likely to ignore those entities with less historical interaction and predict the wrong entities that frequently interact with other events. The proposed historical contrastive learning trains contrastive representations of queries to identify a small number of highly correlated entities at the query level.

Specifically, the training process of supervised contrastive learning~\cite{khosla2020supervised} consists of two stages. We first introduce $I_q$ to indicate whether the missing object is in $\mathcal{H}_{t}^{s,p}$ for query $q$. In other words, if $I_q$ is equal to $1$, the missing object of the given query $q$ is in $\mathcal{H}_{t}^{s,p}$, and $0$ otherwise. The aim of the two stages is to train a binary classifier which infers the value of such boolean scalar for query $q$.

\textbf{Stage 1: Learning Contrastive Representations.}
In the first stage, the model learns the contrastive representations of queries by minimizing supervised contrastive loss, which takes whether $I_q$ is positive as the training criterion to separate representations of different queries as far as possible in semantic space. Let $\textbf{v}_q$ be the embedding vector (representation) of the given query $q$:

\begin{equation}\label{eq:v_query}
    \textbf{v}_q = MLP(\textbf{s} \oplus \textbf{p} \oplus tanh(\textbf{W}_{F}\textbf{F}_{t}^{s,p})),
\end{equation}
where the query's information is encoded by an MLP to normalize and project the embedding onto the unit sphere for further contrastive training. Let $M$ denote the minibatch, $Q(q)$ denote the set of queries in the $M$ except $q$ whose boolean labels are the same as $I_q$, given as: 

\begin{equation}
    Q(q) = \bigcup_{m \in M \backslash \{q\}}\{m|I_m = I_q\}.
\end{equation}

The detail of computing supervised contrastive loss $\mathcal{L}^{sup}$ in the first stage is as follows:

\begin{equation}\label{eq:sup_loss}
    \mathcal{L}^{sup} = \sum_{q \in M}\frac{-1}{|Q(q)|}\sum_{k \in Q(q)}\log \frac{exp(\textbf{v}_q \cdot \textbf{v}_k / \tau)}{\sum\limits_{a \in M \backslash \{q\}}(\textbf{v}_q \cdot \textbf{v}_a / \tau)},
\end{equation}
where, $\textbf{W}_{F} \in \mathbb{R}^{d \times |\mathcal{E}|}$ is the trainable parameter, $\tau \in \mathbb{R}^{+}$ is the temperature parameter set to 0.1 in experiments as recommended in the previous work~\cite{khosla2020supervised}. The objective of $\mathcal{L}^{sup}$ is to make the representations of the same category closer. It should be noted that the contrastive supervised loss $\mathcal{L}^{sup}$ and the previous cross-entropy-like loss $\mathcal{L}^{ce}$ are trained simultaneously.

\textbf{Stage 2: Training Binary Classifier.}
When the training of the first stage is finished, CENET freezes the weights of corresponding parameters including $\textbf{E}$, $\textbf{P}$ and their encoders in the first stage. Then it feeds $\textbf{v}_q$ to a linear layer to train a binary classifier with cross-entropy loss according to the ground truth $I_q$, which is trivial to mention. Now, the classifier can recognize whether the missing object entity of query $q$ exists in the set of historical entities. 

In the process of reasoning, a boolean mask vector $\textbf{B}_t^{s,p} \in \mathbb{R}^{|\mathcal{E}|}$ is generated to identify which kind of entities should be concerned according to the predicted $\hat{I}_q$ and whether $o \in \mathcal{H}_{t}^{s,p}$ is true:

\begin{equation}\label{eq:mask}
    \textbf{B}_t^{s,p}(o) = \Phi_{o \in \mathcal{H}_{t}^{s,p} = \hat{I}_q}.
\end{equation}
The probabilities of entities in all positive positions ($\textbf{B}_t^{s,p}(o)=1$) will be further increased, and vice versa. In other words, if the missing object is predicted to be in $\mathcal{H}_{t}^{s,p}$, then entities in the historical set will receive more attention. Otherwise, those entities outside the historical set are more likely to be attended.




\subsection{Parameter Learning and Inference}
We minimize the loss function in the first stage:
\begin{equation}\label{eq:total_loss}
    \mathcal{L}=\alpha \cdot \mathcal{L}^{ce} + (1-\alpha) \cdot \mathcal{L}^{sup},
\end{equation}
where $\alpha$ is a hyper-parameter ranging from 0 to 1 to balance different losses. As to the second stage, we choose binary cross-entropy with sigmoid activation to train the binary classifier. Taking the prediction of object entities as an example, the detailed training process of CENET is provided in Algorithm~\ref{alg:training} (See Appendix 2 for the computational complexity). Such a training process is also used to predict the missing subject entities in the experiments.


\begin{algorithm}[tb]
\caption{Learning algorithm of CENET}
\label{alg:training}
{\bf Input:} Observed graph quadruples set $\mathcal{G}$, entity set $\mathcal{E}$, relation type set $\mathcal{R}$, hyper-paratermeter $\alpha$, and $\lambda$.\\
{\bf Output:} A trained network.
\begin{algorithmic}[1] 
\STATE Initiate parameters of network $Net$;
\FOR{each $(s,p,o,t)$ in $\mathcal{G}$}
\STATE Compute $\mathcal{H}_{t}^{s,p}$, $\textbf{F}_{t}^{s,p}$, and $\textbf{Z}_{t}^{s,p}$ for query $(s,p,?,t)$ according to Eq.\ref{eq:historical_entity}, \ref{eq:frequency}, and \ref{eq:frequency_trans} respectively;
\STATE Label $I_q$ for query $(s,p,?,t)$ using $\mathcal{H}_{t}^{s,p}$;
\ENDFOR
\WHILE{loss does not converge}
\STATE Compute $\textbf{H}_{his}^{s,p}$ and $\textbf{H}_{nhis}^{s,p}$ using $\textbf{Z}_{t}^{s,p}$ according to Eq.\ref{eq:hhis} and \ref{eq:hnhis} for each $(s,p,o,t)$ in $\mathcal{G}$;
\STATE Compute $\textbf{v}_q$ using $\textbf{F}_{t}^{s,p}$ according to Eq.\ref{eq:v_query};
\STATE Compute $\mathcal{L}^{ce}$ using $\textbf{H}_{his}^{s,p}$ and $\textbf{H}_{nhis}^{s,p}$ according to Eq.\ref{eq:ce_loss};
\STATE Compute $\mathcal{L}^{sup}$ using $\textbf{v}_q$ according to Eq.\ref{eq:sup_loss};
\STATE $\mathcal{L} \leftarrow \alpha \cdot \mathcal{L}^{ce} + (1-\alpha) \cdot \mathcal{L}^{sup}$;
\STATE Optimize $Net$ according to $\mathcal{L}$;
\ENDWHILE
\STATE Freeze parameters of $Net$ except the classification layer in the second stage;
\STATE Train the classification layer in $Net$ according to $I_q$ and $\textbf{v}_q$ with binary cross-entropy;
\STATE \textbf{return} $Net$;
\end{algorithmic}
\end{algorithm}

As can be seen from Figure~\ref{fig:architecture}, the middle part illustrates the inference process that receives the distribution $\textbf{P}^{s,p}_{t}$ and the mask vector $\textbf{B}_t^{s,p}$ from both sides respectively. Then, CENET will choose the object with the highest probability as the final prediction $\hat{o}$:

\begin{equation}\label{eq:mask_hard}
    \textbf{P}(o|s,p,\textbf{F}_{t}^{s,p}) = \textbf{P}^{s,p}_{t}(o) \cdot \textbf{B}_t^{s,p}(o),
\end{equation}

\begin{equation}\label{eq:o_hat}
    \hat{o} = argmax_{o \in \mathcal{E}}\textbf{P}(o|s,p,\textbf{F}_{t}^{s,p}).
\end{equation}
Additionally, it is possible that a poor classifier of the second stage of historical contrastive learning may deteriorate the performance when wrongly masking the expected object entities. Thus, there is a compromised substitution:
\begin{equation}\label{eq:mask_soft}
    \textbf{P}(o|s,p,\textbf{F}_{t}^{s,p}) = \textbf{P}^{s,p}_{t}(o) \cdot softmax(\textbf{B}_t^{s,p})(o).
\end{equation}
We call the former version in Equation~\ref{eq:mask_hard} \textit{hard-mask}, the latter in Equation~\ref{eq:mask_soft} \textit{soft-mask}. The hard-mask can reduce the search space and the soft-mask can obtain a more convincing distribution which makes the model more conservative. 


\section{Experiments}\label{sec:exp}
This section conducts a series of experiments to validate the performance of CENET. We first present the experimental settings and then compare CENET with a wide selection of TKG models. After that, the ablation study is implemented to evaluate the effectiveness of various components. Finally, the analysis of hyper-parameter is discussed. All our datasets and codes are publicly available\footnote{https://github.com/xyjigsaw/CENET}.


\subsection{Experimental Settings}
\subsubsection{\textbf{Datasets and Baselines}}
We select five benchmark datasets, including three event-based TKGs and two public KGs. These two types of datasets are constructed in different ways. The former three event-based TKGs consist of \textit{Integrated Crisis Early Warning System} (ICEWS18~\cite{boschee2015icews} and ICEWS14~\cite{trivedi2017know}) and \textit{Global Database of Events, Language, and Tone} (GDELT~\cite{leetaru2013gdelt}) where a single event may happen at any time. The last two public KGs (WIKI~\cite{leblay2018deriving} and YAGO~\cite{mahdisoltani2014yago3}) consist of temporally associated facts which last a long time and hardly occur in the future. Table~\ref{tab:datasets} provides the statistics of these datasets.


\begin{table}[ht]
\centering
\tabcolsep 0.015in
\begin{tabular}{cccccc}
\hline
\textbf{Dataset} & \multicolumn{1}{c}{\textbf{Entities}} & \multicolumn{1}{c}{\textbf{Relation}} & \multicolumn{1}{c}{\textbf{Training}} & \multicolumn{1}{c}{\textbf{Validation}} & \multicolumn{1}{c}{\textbf{Test}}  \\ \hline
ICEWS18          & 23,033                                & 256                                   & 373,018                               & 45,995                                  & 49,545                                              \\
ICEWS14          & 12,498                                & 260                                   & 323,895                               & -                                       & 341,409                                             \\
GDELT            & 7,691                                 & 240                                   & 1,734,399                             & 238,765                                 & 305,241                                              \\
WIKI             & 12,554                                & 24                                    & 539,286                               & 67,538                                  & 63,110                                                 \\
YAGO             & 10,623                                & 10                                    & 161,540                               & 19,523                                  & 20,026                                                   \\ \hline
\end{tabular}
\caption{Statistics of the datasets.}
\label{tab:datasets}
\end{table}

\begin{table*}[t]
\small
\centering
\tabcolsep 0.061in
\begin{tabular}{c|cccc|cccc|cccc}
\toprule
\multirow{2}{*}{Method} & \multicolumn{4}{c|}{ICEWS18}                                      & \multicolumn{4}{c|}{ICEWS14}                                      & \multicolumn{4}{c}{GDELT}                     \\ \cmidrule{2-13} 
                        & MRR            & Hits@1         & Hits@3         & Hits@10        & MRR            & Hits@1         & Hits@3         & Hits@10        & MRR       & Hits@1    & Hits@3    & Hits@10   \\ \midrule
TransE                  & 17.56          & 2.48           & 26.95          & 43.87          & 18.65          & 1.12           & 31.34          & 47.07          & 16.05     & 0.00      & 26.10     & 42.29     \\
DistMult                & 22.16          & 12.13          & 26.00          & 42.18          & 19.06          & 10.09          & 22.00          & 36.41          & 18.71     & 11.59     & 20.05     & 32.55     \\
ComplEx                 & 30.09          & 21.88          & 34.15          & 45.96          & 24.47          & 16.13          & 27.49          & 41.09          & 22.77     & 15.77     & 24.05     & 36.33     \\
R-GCN                   & 23.19          & 16.36          & 25.34          & 36.48          & 26.31          & 18.23          & 30.43          & 45.34          & 23.31     & 17.24     & 24.96     & 34.36     \\
ConvE                   & 36.67          & 28.51          & 39.80          & 50.69          & 40.73          & 33.20          & 43.92          & 54.35          & 35.99     & 27.05     & 39.32     & 49.44     \\ \midrule
TeMP                    & 40.48          & 33.97          & 42.63          & 52.38          & 43.13          & 35.67          & 45.79          & 56.12          & 37.56     & 29.82     & 40.15     & 48.60     \\
RE-NET                  & 42.93          & 36.19          & 45.47          & 55.80          & 45.71          & 38.42          & 49.06          & 59.12          & 40.12     & 32.43     & 43.40     & 53.80     \\
xERTE                   & 36.95          & 30.71          & 40.38          & 49.76          & 32.92          & 26.44          & 36.58          & 46.05          &           &\quad\quad$\gg$  & 1 day\quad\quad     &      \\
TLogic                  & 37.52         & 30.09          & 40.87          & 52.27          & 38.19          & 32.23          & 41.05          & 49.58          & 22.73      & 17.65     & 24.66     & 32.59    \\
RE-GCN                   & 32.78         & 24.99          & 35.54          & 48.01          & 32.37          & 24.43          & 35.05          & 48.12          & 29.46      & 21.74     & 32.01     & 43.62    \\
TANGO-TuckER            & 44.56          & 37.87          & 47.46          & 57.06          & 46.42          & 38.94          & 50.25          & 59.80          & 38.00     & 28.02     & 43.91     & 53.70     \\
TANGO-Distmult          & 44.00          & 38.64          & 45.78          & 54.27          & 46.68          & 41.20          & 48.64          & 57.05          & 41.16     & 35.11     & 43.02     & 52.58     \\
CyGNet                  & 46.69          & 40.58          & 49.82          & 57.14          & 48.63          & 41.77          & 52.50          & 60.29          & 50.29     & 44.53     & 54.69     & 60.99     \\
EvoKG                   & 29.67          & 12.92          & 33.08          & 58.32          & 18.30          & 6.30          & 19.43          & 39.37          & 11.29     & 2.93     & 10.84     & 25.44     \\
HIP                     & \underline{48.37}          & \underline{43.51}          & \underline{51.32}          & \underline{58.49}          & \underline{50.57}          & \underline{45.73}          & \underline{\textbf{54.28}}          & \underline{\textbf{61.65}}          & \underline{52.76}     & \underline{46.35}     & \underline{55.31}     & \underline{61.87}     \\
\midrule
\textbf{CENET}          & \textbf{51.06} & \textbf{47.10} & \textbf{51.92} & \textbf{58.82} & \textbf{53.35} & \textbf{49.61} & 54.07 & 60.62 & \textbf{58.48} & \textbf{55.99} & \textbf{58.63} & \textbf{62.96} \\ 
\bottomrule
\end{tabular}
\caption{Experimental results of temporal link prediction on three event-based TKGs. $\gg$ \textit{1 day} means running time is more than 1 day. The best results are boldfaced, and the results of previous SOTAs are underlined.}

\label{tab:results3tkg}
\end{table*}


CENET is compared with $15$ up-to-date knowledge graph reasoning models, including static and temporal approaches. Static methods include TransE~\cite{bordes2013translating}, DistMult~\cite{yang2014embedding}, ComplEx~\cite{trouillon2016complex}, R-GCN~\cite{schlichtkrull2018modeling}, and ConvE~\cite{dettmers2018convolutional}. Temporal models include TeMP~\cite{wu2020temp}, RE-NET~\cite{jin2020recurrent}, xERTE~\cite{han2020explainable}, TLogic~\cite{liu2021tlogic}, RE-GCN~\cite{li2021temporal}, TANGO-TuckER~\cite{han2021learning}, TANGO-Distmult~\cite{han2021learning}, CyGNet~\cite{zhu2021learning}, EvoKG~\cite{park2022evokg}, and HIP~\cite{he2021hip}. 


\subsubsection{\textbf{Training Settings and Evaluation Metrics}}
All datasets except ICEWS14 are split into training set (80\%), validation set (10\%), and testing set (10\%). The original ICEWS14 is not provided with a validation set. We report a widely used filtered version~\cite{jin2020recurrent,han2020explainable,zhu2021learning,he2021hip} of Mean Reciprocal Ranks (MRR) and Hits@1/3/10 (the proportion of correct predictions ranked within top 1/3/10). As to model configurations, we set the batch size to 1024, embedding dimension to 200, learning rate to 0.001, and use Adam optimizer. The training epoch for $\mathcal{L}$ is limited to 30, and the epoch for the second stage of contrastive learning is limited to 20. The value of hyper-parameter $\alpha$ is set to 0.2, and $\lambda$ is set to 2. For the settings of baselines, we use their recommended configurations. 


\subsection{Results}

\subsubsection{\textbf{Results on Event-based TKGs}}
Table~\ref{tab:results3tkg} presents the MRR and Hits@1/3/10 results of link (event) prediction on three event-based TKGs. Our proposed CENET outperforms other baselines in most cases. It can be observed that many static models are inferior to temporal models because static models do not consider temporal information and their dependency between different snapshots. In the case of temporal models, TeMP is designed to complete missing links (graph interpolation) rather than predict new events, and it thus shows worse performance than extrapolation models. Although xERTE provides a certain degree of predictive explainability, it is computationally inefficient to handle large-scale datasets such as GDELT, whose training set contains more than $1$ million samples. In terms of Hits@10, CENET is on par with HIP on these three event-based datasets. Nevertheless, the results of Hits@1 improve the most in our model. CENET achieves up to \textbf{8.25\%, 8.48\%, and 20.80\%} improvements of Hits@1 on ICEWS18, ICEWS14, and GDELT respectively. The main reason is that there exist a large proportion of new events without historical events in event-based datasets. CENET learns the historical and non-historical dependency of new events simultaneously, which mines those unobserved underlying factors. In contrast, models including TANGO and HIP perform well in terms of Hits@10 but cannot predict the correct entities exactly, making Hits@1 much lower than ours.

\begin{table}[ht]
\small
\centering
\tabcolsep 0.018in
\begin{tabular}{c|ccc|ccc}
\toprule
\multirow{2}{*}{Method} & \multicolumn{3}{c|}{WIKI}                        & \multicolumn{3}{c}{YAGO}                        \\ \cmidrule{2-7} 
                        & MRR            & Hits@1         & Hits@3                & MRR            & Hits@1            & Hits@3     \\ \midrule
TransE                  & 46.68          & 36.19          & 49.71                 & 48.97          & 46.23          & 62.45                    \\
DistMult                & 46.12          & 37.24          & 49.81                 & 59.47          & 52.97          & 60.91                    \\
ComplEx                 & 47.84          & 38.15          & 50.08                 & 61.29          & 54.88          & 62.28                    \\
R-GCN                   & 37.57          & 28.15          & 39.66                 & 41.30          & 32.56          & 44.44                    \\
ConvE                   & 47.57          & 38.76          & 50.10                 & 62.32          & 56.19          & 63.97                    \\ \midrule
TeMP                    & 49.61          & 46.96          & 50.24                 & 62.25          & 55.39          & 64.63                    \\
RE-NET                  & 51.97          & 48.01          & 52.07                 & 65.16          & 63.29          & 65.63                   \\
xERTE                   &           & $\gg$ 1 day  &              & 58.75          & 58.46          & 58.85                  \\
TLogic                  & \underline{57.73}& \underline{57.43}&57.88  & 1.29          & 0.49          & 0.85                 \\
RE-GCN                  & 44.86          & 39.82          & 46.75                 & 65.69          & 59.98          & 68.70                 \\
TANGO-TuckER            & 53.28          & 52.21          & 53.61                 & 67.21          & 65.56          & 67.59                  \\
TANGO-Distmult          & 54.05          & 51.52          & 53.84                 & \underline{68.34} & \underline{67.05} & 68.39             \\
CyGNet                  & 45.50          & 50.48          & 50.79                 & 63.47          & 64.26          & 65.71                  \\ 

EvoKG                   & 50.66          & 12.21          & \underline{63.84}                 & 55.11          & 54.37          & \underline{81.38}                  \\ 

HIP                     & 54.71          & 53.82    & 54.73                       & 67.55  & 66.32          & 68.49                \\
\midrule
\textbf{CENET}          & \textbf{68.39} & \textbf{68.33} & \textbf{68.36} & \textbf{84.13} & \textbf{84.03} & \textbf{84.23}  \\ \bottomrule
\end{tabular}
\caption{Experimental results of temporal link prediction on two public KGs. See Appendix for more results.}
\vspace{-0.6cm}
\label{tab:results2tkg}
\end{table}

\begin{table*}[t]
\small
\centering
\tabcolsep 0.14in
\scalebox{0.98}{
\begin{tabular}{l|ccll|ccll}
\toprule
\multicolumn{1}{c|}{\multirow{2}{*}{Method}} & \multicolumn{4}{c|}{ICEWS18}                                             & \multicolumn{4}{c}{YAGO}                                                 \\ \cmidrule{2-9} 
\multicolumn{1}{c|}{}                        & MRR                       & Hits@1                    & Hits@3 & Hits@10 & MRR                       & Hits@1                    & Hits@3 & Hits@10 \\ \midrule
CENET-his                                 & 50.65                     & \textbf{47.15}                     & 51.23  & 57.42   & 71.64                    & 70.24                      & 71.81  & 74.39    \\
CENET-nhis                                & 31.75                     & 24.22                     & 34.01  & 46.69   & 61.73                    & 59.64                      & 62.50  & 65.38    \\

CENET-$\mathcal{L}^{ce}$ (w/o-stage-1)         & 50.59                     & 46.47                     & 51.58  & 58.58   & 75.25                     & 73.96                     & 75.55  & 77.52   \\
CENET-w/o-stage-2                  & 50.32                     & 46.30                     & 51.29  & 58.16   & 77.53                     & 76.12                     & 78.04  & 79.84   \\ 
CENET-w/o-CL       & 49.98                     & 45.89                     & 50.74  & 57.81   & 73.29                     & 71.87                     & 73.64  & 75.90   \\
CENET-random-mask                       & 26.80                     & 24.42                     & 27.47  & 31.60   & 39.07                     & 38.31                     & 39.28  & 40.41   \\ 
CENET-hard-mask                         & 49.66                     & 46.69                     & 50.78  & 55.75   & \textbf{84.13}                     & \textbf{84.03}                     & \textbf{84.23}  & \textbf{84.24}   \\
CENET-soft-mask                   & \textbf{51.06}            & 47.10            & \textbf{51.92}  & \textbf{58.82}   & 80.03   & 79.09   & 80.30  & 81.57   \\ \midrule
CENET-GT-mask                           & 52.75                     & 48.21                     & 53.97  & 61.84   & 84.73                     & 84.31                     & 84.76  & 85.34   \\ \bottomrule
\end{tabular}}
\caption{Ablation study of CENET on ICEWS18 and YAGO.}
\vspace{-0.4cm}
\label{tab:ablation}

\end{table*}

\subsubsection{\textbf{Results on Public KGs}}
CENET also outperforms the baselines in all metrics on WIKI and YAGO. As can be seen from Table~\ref{tab:results2tkg}, CENET significantly achieves the improvements up to \textbf{23.68\% (MRR), 25.77\% (Hits@1), and 7.08\% (Hits@3)} over SOTA on public KGs. This is because the recurrence rates in these two datasets are imbalanced~\cite{zhu2021learning}, and our model can easily handle such data. In terms of the WIKI dataset, $62.3\%$ object entities associated with their corresponding facts (grouped by \textit{(subject, relation)} tuples) have appeared repeatedly at least once in history. In contrast, the recurrence rate of subject entities (grouped by \textit{(object, relation)} tuples) is $23.4\%$, which hinders many models learning from the historical information when inferring subject entities. CENET can effectively alleviate the problem of the imbalanced recurrence rate because the concurrent learning of historical and non-historical dependency can complement each other to generate the entity distribution. Also, the probability of selecting unrelated entities is greatly reduced on account of the binary classifier regardless of the imbalanced recurrence rate.



\subsection{Ablation Study}\label{ablation}
We choose ICEWS18 and YAGO to investigate the effectiveness of the historical/non-historical dependency, contrastive learning, and the mask-based inference. Table~\ref{tab:ablation} shows the results of ablation.

CENET-his only considers the historical dependency while CENET-nhis keeps the non-historical dependency. Both of them employ the contrastive learning. The performance of CENET-his is better than CENET-nhis since most events can be traced to their historical events especially in event-based TKGs. Still, for CENET-nhis, it also works on event prediction to a certain extent. Thus, it is necessary to consider both dependencies at the same time. We remove $\mathcal{L}^{sup}$ and only retain $\mathcal{L}^{ce}$ as the variant CENET-$\mathcal{L}^{ce}$. In the case of ICEWS18, the $\mathcal{L}^{ce}$ is capable of achieving high results close to the proposed CENET, while the results in YAGO have \textbf{dropped about 7\%}. Such results verify the positive influence of the stage 1 in the historical contrastive learning. CENET-w/o-stage-2 is another variant that minimizes $\mathcal{L}^{ce}$ and $\mathcal{L}^{sup}$ without training the binary classifier, which naturally discards the mask-based inference. Such changes cause \textbf{1.7\% and 3.8\% drop} in terms of Hits@1 on ICEWS18 and YAGO respectively. CENET-w/o-CL removing the historical contrastive learning has worse performance than the above two variants. These results prove the significance of our proposed historical contrastive learning. As to the mask strategy. The mask vector is a randomly generated boolean vector in CENET-random-mask. CENET-hard-mask and CENET-soft-mask are our proposed two ways to tackle the mask vector. We use the ground truth in the testing set to generate a mask vector represented by CENET-GT-mask to explore the upper bound of CENET. We can see that untrained model with randomly generated mask vector is counterproductive to the prediction.


\begin{figure}[htb]
    \centering
    \includegraphics[width=0.95\linewidth]{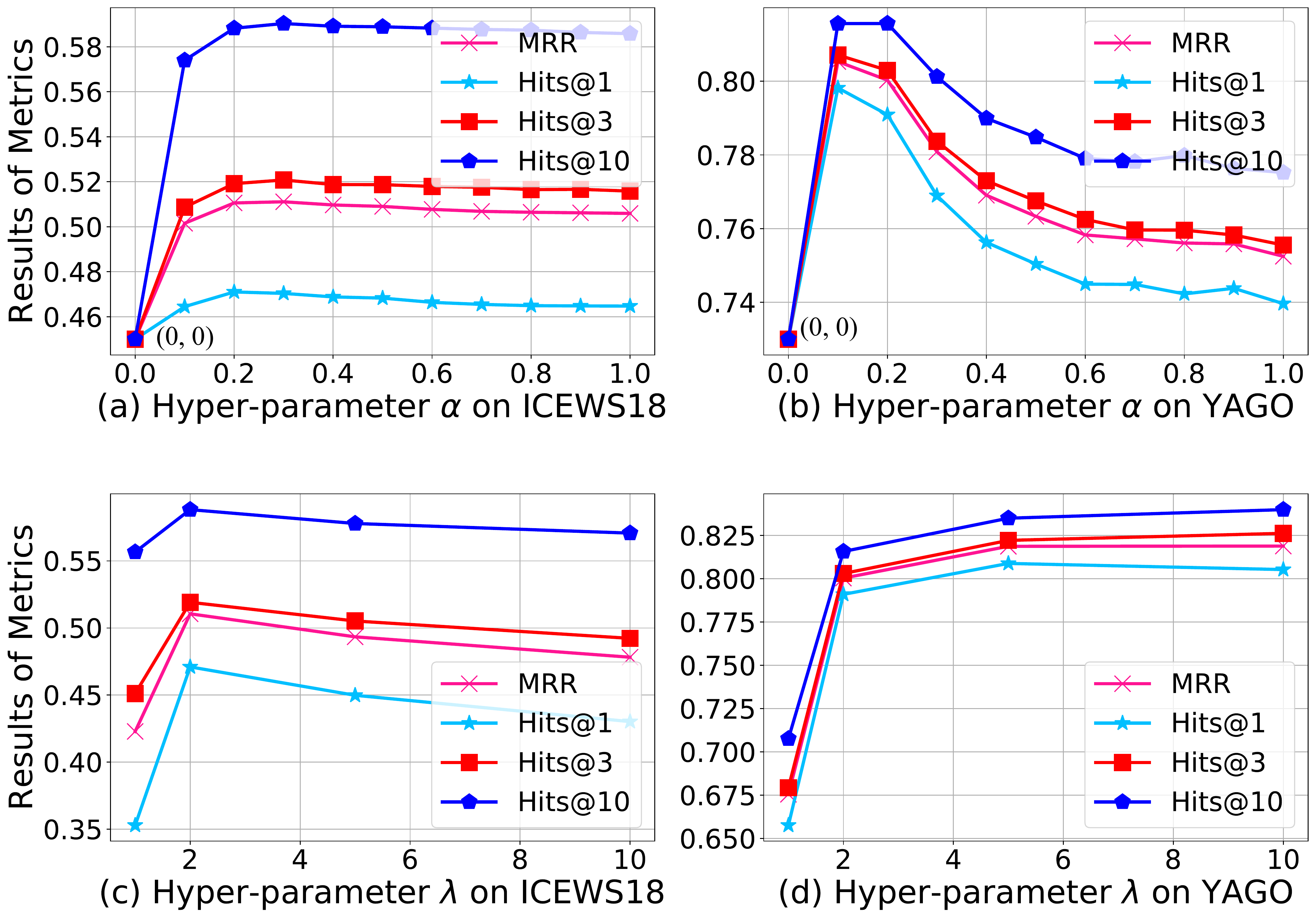}
    \caption{Results of hyper-parameters $\alpha$ and $\lambda$ of CENET on ICEWS18 and YAGO.}
    \vspace{-0.2cm}
    \label{fig:hyperparameter}
\end{figure}

\subsection{Hyper-parameter Analysis}
There are two unexplored hyper-parameters $\alpha$ and $\lambda$ in CENET. We adjust the values of $\alpha$ and $\lambda$ respectively to observe the performance change of CENET on ICEWS18 and YAGO. The results are shown in Figure~\ref{fig:hyperparameter}. The hyper-parameter $\alpha$ aims at balancing the contribution of $\mathcal{L}^{ce}$ and $\mathcal{L}^{sup}$. Due to the difference of characteristics between event-based TKGs and public KGs, the hyper-parameter $\alpha$ ranging from 0 to 1 leads to different results on these two kinds of datasets. Specifically, $\mathcal{L}^{ce}$ contributes more to event-based TKGs, while $\mathcal{L}^{sup}$ is more friendly to public KGs. Considering that if we remove $\mathcal{L}^{ce}$ i.e. set $\alpha$ to 0, then we cannot obtain the final probability $\textbf{P}(o|s,p,\textbf{F}_{t}^{s,p})$ (and $\textbf{P}(s|o,p,\textbf{F}_{t}^{o,p})$) for inference. To this end, we set $\alpha$ to 0.2. With regard to the hyper-parameter $\lambda$, we first fix the value of hyper-parameter $\alpha$, then the $\lambda$ is analyzed. We can see that the higher the value of $\lambda$, the better the result on YAGO, whereas the worse the result on ICEWS18. Therefore, $\lambda$ is set to 2.

\section{Conclusion and Future Work}\label{sec:conclusion}
In this paper, we propose a novel temporal knowledge graph representation learning model, Contrastive Event Network (CENET), for event forecasting. The key idea of CENET is to learn a convincing distribution of the whole entity set and identify significant entities from both historical and non-historical dependency in the framework of contrastive learning. The experimental results present that CENET outperforms all existing methods in most metrics significantly, especially for Hits@1. Promising future work includes exploring the ability of contrastive learning in knowledge graph, such as finding more reasonable contrastive pairs.

\section*{Acknowledgments}
This work was supported by NSF China (No. 62020106005, 42050105, 62061146002, 61960206002), 100-Talents Program of Xinhua News Agency, Shanghai Pilot Program for Basic Research - Shanghai Jiao Tong University, and the Program of Shanghai Academic/Technology Research Leader under Grant No. 18XD1401800.

\bibliography{aaai23}

\clearpage

\appendix

\section{Historical Contrastive Learning}
\begin{figure}[ht]
    \centering
    \includegraphics[width=1.0\linewidth]{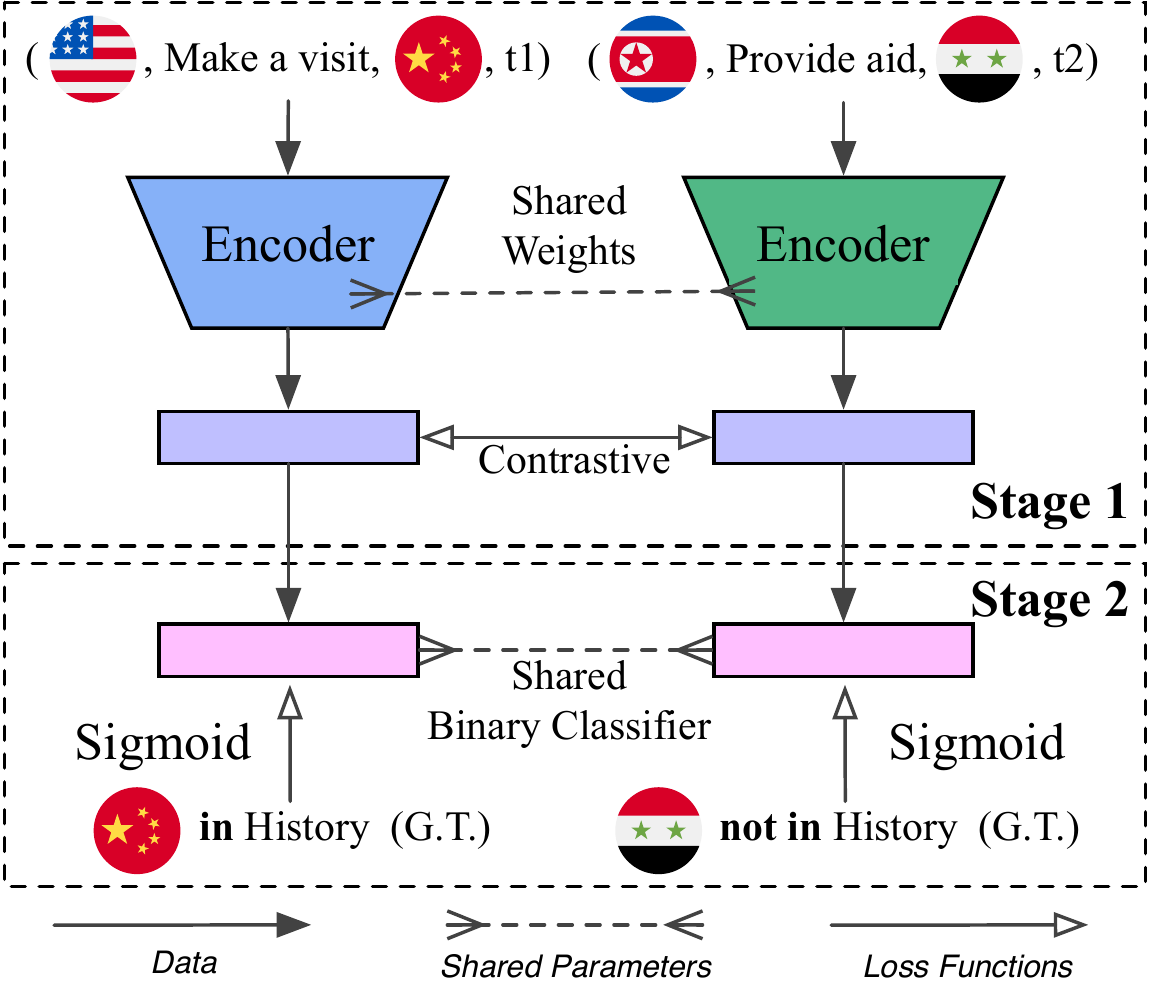}
    \caption{The detail of historical contrastive learning: CENET learns representations using a contrastive loss in stage 1, then trains a binary classifier using cross-entropy loss in stage 2.}
    \label{fig:contrastive_loss}
\end{figure}

The historical contrastive learning consists of two stages: learning contrastive representations and training binary classifier. Figure~\ref{fig:contrastive_loss} illustrates the training process of historical contrastive learning.

\section{Complexity Analysis}
We analyze the complexity of the whole framework of historical contrastive learning in the stage of training and inference. For a minibatch $M$, the time complexity of learning the historical and non-historical dependency is $O(|M||\mathcal{E}|)$. Training the contrastive representation has the time complexity of $O(|M|^2)$ to calculate the similarity matrix for supervised contrastive loss. Thus, the computational complexity of training is $O(|M||\mathcal{E}| + |M|^2)$, and the inference complexity for a single query is $O(|\mathcal{E}|)$ since we only need to calculate the final distribution. The total space complexity is $O(|\mathcal{R}| + |\mathcal{E}| + L)$, where $L$ is the number of layers of neural modules.

\begin{table*}[ht]
\centering
\tabcolsep 0.03in
\begin{tabular}{ccccccccc}
\hline
\textbf{Dataset} & \multicolumn{1}{c}{\textbf{Entities}} & \multicolumn{1}{c}{\textbf{Relation}} & \multicolumn{1}{c}{\textbf{Training}} & \multicolumn{1}{c}{\textbf{Validation}} & \multicolumn{1}{c}{\textbf{Test}} & \multicolumn{1}{c}{\textbf{Granularity}} & \multicolumn{1}{c}{\textbf{Time Granules}} & \multicolumn{1}{c}{\textbf{Proportion of New Events}} \\ \hline
ICEWS18          & 23,033                                & 256                                   & 373,018                               & 45,995                                  & 49,545                            & 24 hours                                 & 304                                        & 39.4\%                                                  \\
ICEWS14          & 12,498                                & 260                                   & 323,895                               & -                                       & 341,409                           & 24 hours                                 & 365                                        & 32.5\%                                                  \\
GDELT            & 7,691                                 & 240                                   & 1,734,399                             & 238,765                                 & 305,241                           & 15 mins                                  & 2,751                                      & 17.8\%                                                  \\
WIKI             & 12,554                                & 24                                    & 539,286                               & 67,538                                  & 63,110                            & 1 year                                   & 232                                        & 2.8\%                                                   \\
YAGO             & 10,623                                & 10                                    & 161,540                               & 19,523                                  & 20,026                            & 1 year                                   & 189                                        & 10.3\%                                                  \\ \hline
\end{tabular}
\caption{Statistics of the datasets.}
\label{tab:datasets-appendix}
\end{table*}

\begin{table*}[ht]
\small
\centering
\tabcolsep 0.055in
\begin{tabular}{c|cccc|cccc|cccc}
\hline
\multirow{2}{*}{Method} & \multicolumn{4}{c|}{ICEWS18}                                      & \multicolumn{4}{c|}{ICEWS14}                                      & \multicolumn{4}{c}{GDELT}                     \\ \cline{2-13} 
                        & MRR            & Hits@1         & Hits@3         & Hits@10        & MRR            & Hits@1         & Hits@3         & Hits@10        & MRR       & Hits@1    & Hits@3    & Hits@10   \\ \hline
RotatE                  & 23.10          & 14.33          & 27.61          & 38.72          & 29.56          & 22.14          & 32.92          & 42.68          & 22.33     & 16.68     & 23.89     & 32.29     \\
CompGCN                 & 23.31          & 16.52          & 25.37          & 36.61          & 26.46          & 18.38          & 30.64          & 45.61          & 23.46     & 16.65     & 25.54     & 34.58     \\
HyTE                    & 7.31           & 3.10           & 7.50           & 14.95          & 11.48          & 5.64           & 13.04          & 22.51          & 6.37      & 0.00      & 6.72      & 18.63     \\
TTransE                 & 8.36           & 1.94           & 8.71           & 21.93          & 6.35           & 1.23           & 5.80           & 16.65          & 5.52      & 0.47      & 5.01      & 15.27     \\
TA-DistMult             & 28.53          & 20.30          & 31.57          & 44.96          & 20.78          & 13.43          & 22.80          & 35.26          & 29.35     & 22.11     & 31.56     & 41.39     \\
DySAT                   & 19.95          & 14.42          & 23.67          & 26.67          & 18.74          & 12.23          & 19.65          & 21.17          & 23.34     & 14.96     & 22.57     & 27.83     \\
Know-Evolve+MLP         & 9.29           & 5.11           & 9.62           & 17.18          & 22.89          & 14.31          & 26.68          & 38.57          & 22.78     & 15.40     & 25.49     & 35.41     \\
DyRep+MLP               & 9.86           & 5.14           & 10.66          & 18.66          & 24.61          & 15.88          & 28.87          & 39.34          & 23.94     & 15.57     & 27.88     & 36.58     \\
R-GCRN+MLP              & 35.12          & 27.19          & 38.26          & 50.49          & 36.77          & 28.63          & 40.15          & 52.33          & 37.29     & 29.00     & 41.08     & 51.88     \\
\hline
\textbf{CENET}          & \textbf{51.06} & \textbf{47.10} & \textbf{51.92} & \textbf{58.82} & \textbf{53.35} & \textbf{49.61} & \textbf{54.07} & \textbf{60.62} & \textbf{58.48} & \textbf{55.99} & \textbf{58.63} & \textbf{62.96} \\ \hline
\end{tabular}
\caption{Experimental results of temporal link prediction on three event-based TKGs (ICEWS18, ICEWS14, GDELT). $\gg$ \textit{1 day} means running time is more than 1 day. The best results are boldfaced.}
\label{tab:results3tkg-appendix}
\end{table*}

\begin{table*}[ht]
\small
\tabcolsep 0.13in
\centering
\begin{tabular}{c|cccc|cccc}
\hline
\multirow{2}{*}{Method} & \multicolumn{4}{c|}{WIKI}                        & \multicolumn{4}{c}{YAGO}                        \\ \cline{2-9} 
                        & MRR            & Hits@1         & Hits@3         & Hits@10        & MRR            & Hits@1            & Hits@3         & Hits@10        \\ \hline
RotatE                  & 50.67          & 40.88          & 50.71          & 50.88          & 65.09          & 57.13          & 65.67          & 66.16          \\ 
CompGCN                 & 37.64          & 28.33          & 39.87          & 42.03          & 41.42          & 32.63          & 44.59          & 52.81          \\
HyTE                    & 43.02          & 34.29          & 45.12          & 49.49          & 23.16          & 12.85          & 45.74          & 51.94          \\
TTransE                 & 31.74          & 22.57          & 36.25          & 43.45          & 32.57          & 27.94          & 43.39          & 53.37          \\
TA-DistMult             & 48.09          & 38.71          & 49.51          & 51.70          & 61.72          & 52.98          & 65.32          & 67.19          \\
DySAT                   & 31.82          & 22.07          & 26.59          & 35.59          & 43.43          & 31.87          & 43.67          & 46.49          \\
Know-Evolve+MLP         & 12.64          &   -            & 14.33          & 21.57          & 6.19           &   -            & 6.59           & 11.48          \\
DyRep+MLP               & 11.60          &   -            & 12.74          & 21.65          & 5.87           &   -            & 6.54           & 11.98          \\
R-GCRN+MLP              & 47.71          &   -            & 48.14          & 49.66          & 53.89          &   -            & 56.06          & 61.19          \\ \hline
\textbf{CENET}          & \textbf{68.39} & \textbf{68.33} & \textbf{68.36} & \textbf{68.47} & \textbf{84.13} & \textbf{84.03} & \textbf{84.23} & \textbf{84.24} \\ \hline
\end{tabular}
\caption{Experimental results of temporal link prediction on two public KGs (WIKI and YAGO). The best results are boldfaced.}
\label{tab:results2tkg-appendix}
\end{table*}

\section{Details of Datasets}
We select five benchmark datasets, including three event-based TKGs and two public KGs. These two types of datasets are constructed in different ways. The former three event-based TKGs consist of ICEWS18, ICEWS14, and GDELT, where a single event may happen at any time. The last two public KGs (WIKI and YAGO) consist of temporally associated facts which last a long time and hardly occur in the future. In our experiment, the preprocessing of these datasets is similar to that of RE-NET~\cite{jin2020recurrent}. Table~\ref{tab:datasets-appendix} provides the detailed statistics of these datasets. As mentioned earlier, the number of new events has a great impact on the performance of different models. We provide the rates of new events on different training datasets. It can be observed that there are a large proportion of new events in event-based TKGs (ICEWS18, ICEWS14, and GDELT). In terms of all public KGs (WIKI and YAGO), the rates of new events are lower than $20\%$. Predicting new events is a challenge for many autoregressive models because mining the non-historical dependency is difficult. Our proposed historical contrastive learning addresses the issue to some extent.

\section{More Baseline Results}
CENET is compared with $9$ more models, including static and temporal approaches. Static methods include RotatE~\cite{sun2019rotate}, and CompGCN~\cite{vashishth2020composition}. Temporal models include HyTE~\cite{dasgupta2018hyte}, Know-Evolve~\cite{trivedi2017know}, TTransE~\cite{jiang2016towards}, TA-DistMult~\cite{garcia2018learning}, DySAT~\cite{sankar2020dysat}, DyRep+MLP~\cite{trivedi2019dyrep}, and R-GCRN+MLP~\cite{seo2018structured}. Some of the temporal baselines (HyTE and TTransE) are not applicable to event forecasting since they are proposed to handle graph interpolation, whereas our work focuses on the extrapolation task. Thus, we deal with them in the way of previous work~\cite{jin2020recurrent,zhu2021learning,he2021hip}, which is trivial to mention. Table~\ref{tab:results3tkg-appendix} (previous page) presents the MRR and Hits@1/3/10 results of link (event) prediction on three event-based TKGs (ICEWS18, ICEWS14, and GDELT). Table~\ref{tab:results2tkg-appendix} shows the results on two public KGs (WIKI and YAGO).

\begin{figure*}[ht]
    \centering
    \includegraphics[width=0.9\linewidth]{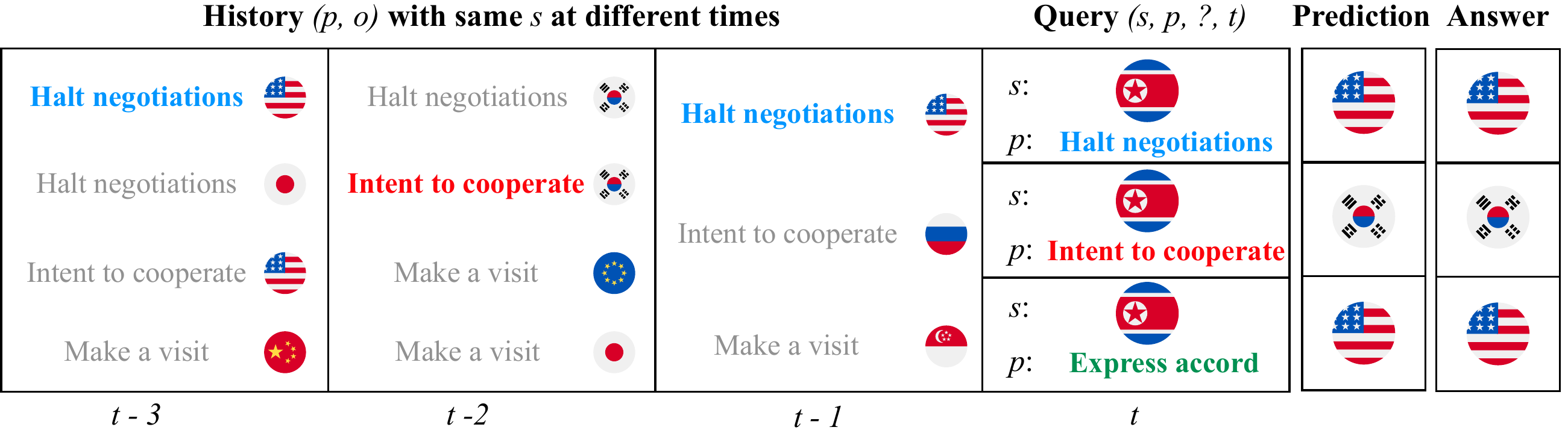}
    \caption{Case study of CENET’s predictions. We select three queries with \textit{North Korea} as subject entity for analysis.}
    \label{fig:case_study}
\end{figure*}

\section{Case Study}

As shown in Figure~\ref{fig:case_study}, we select three representative queries with \textit{North Korea} as subject entity to investigate the predicted results of CENET.

\begin{itemize}
    \item Query \textit{(North Korea, Halt negotiations, ?, t)}: It can be observed that the group \textit{(Halt negotiations, the United States)} appears most frequently in the past (with blue font). It is easy for CENET to obtain the correct answer for the reason that the historical dependency has been captured, and the mask-based inference with a binary classifier can reduce the probabilities of non-historical entities such as \textit{Russia and Singapore} etc, which have nothing to do with the relation `\textit{Halt negotiations}'.
    \item Query \textit{(North Korea, Intent to cooperate, ?, t)}: The group \textit{(Intent to cooperate, South Korea)} only happened once (with red font), which is the same to other object entities such as \textit{the United States} and \textit{Russia}. Not surprisingly, the model can predict correctly. Although the \textit{United States} and \textit{North Korea} had the relation `\textit{Intent to cooperate}' in the past, CENET believes that other relations between the \textit{United States} and \textit{North Korea} were more likely to happen, the first case is the best evidence. Thus, the model chose \textit{South Korea}.
    \item Query \textit{(North Korea, Express accord, ?, t)}: This query has no historical events from the first timestamp 0, but the model also gets the correct result, demonstrating that CENET has learned the non-historical dependency.
\end{itemize}

\end{document}